\documentclass[conference]{IEEEtran}
\IEEEoverridecommandlockouts
\usepackage{cite}
\usepackage{amsmath,amssymb,amsfonts}
\usepackage{algorithmic}
\usepackage{graphicx}
\usepackage{textcomp}
\usepackage{xcolor}
\def\BibTeX{{\rm B\kern-.05em{\sc i\kern-.025em b}\kern-.08em
    T\kern-.1667em\lower.7ex\hbox{E}\kern-.125emX}}
    
\usepackage{multirow}
\usepackage{color}
\def\red#1{\textcolor[rgb]{1,0,0}{#1}}
\def\green#1{\textcolor[rgb]{0,1,0}{#1}}
\def\blue#1{\textcolor[rgb]{0,0,1}{#1}}
\usepackage{booktabs}
\usepackage{array}
\usepackage[colorlinks,linkcolor=red,anchorcolor=blue,citecolor=green]{hyperref}
\makeatletter
\renewcommand{\maketag@@@}[1]{\hbox{\m@th\normalsize\normalfont#1}}%
\makeatother
\usepackage{hyperref}
\usepackage{marvosym}
\DeclareRobustCommand*{\IEEEauthorrefmark}[1]{%
    \raisebox{0pt}[0pt][0pt]{\textsuperscript{\footnotesize\ensuremath{#1}}}}

\begin{document}

\title{TANet: Transformer-based Asymmetric Network for RGB-D Salient Object Detection\\
% {\footnotesize \textsuperscript{*}Note: Sub-titles are not captured in Xplore and
% should not be used}
% \thanks{}
}

\author{
\IEEEauthorblockN{
Chang Liu\IEEEauthorrefmark{1}$^,$\IEEEauthorrefmark{2},
Gang Yang\IEEEauthorrefmark{1}\href{mailto:yanggang@mail.neu.edu.cn}{$^{(\textrm{\Letter})}$},
Shuo Wang\IEEEauthorrefmark{1},
Hangxu Wang\IEEEauthorrefmark{1}$^,$\IEEEauthorrefmark{2},
Yunhua Zhang\IEEEauthorrefmark{1}, and
Yutao Wang\IEEEauthorrefmark{1}}
\IEEEauthorblockA{\IEEEauthorrefmark{1}Northeastern University, Shenyang 110819, China}
\IEEEauthorblockA{\IEEEauthorrefmark{2}DUT Artificial Intelligence Institute, Dalian 116024, China}
% \IEEEauthorblockA{Corresponding Author: Gang Yang \quad Email: yanggang@mail.neu.edu.cn}
}

% \author{
% \IEEEauthorblockN{Chang Liu, Gang Yang\href{mailto:yanggang@mail.neu.edu.cn}{$^{\textrm{\Letter}}$}, Shuo Wang, Hangxu Wang, Yunhua Zhang and Yutao Wang}
% }

\maketitle

\begin{abstract}
Existing RGB-D SOD methods mainly rely on a symmetric two-stream CNN-based network to extract RGB and depth channel features separately. However, there are two problems with the symmetric conventional network structure: first, the ability of CNN in learning global contexts is limited; second, the symmetric two-stream structure ignores the inherent differences between modalities. In this paper, we propose a Transformer-based asymmetric network (TANet) to tackle the issues mentioned above. We employ the powerful feature extraction capability of Transformer (PVTv2) to extract global semantic information from RGB data and design a lightweight CNN backbone (LWDepthNet) to extract spatial structure information from depth data without pre-training. The asymmetric hybrid encoder (AHE) effectively reduces the number of parameters in the model while increasing speed without sacrificing performance. Then, we design a cross-modal feature fusion module (CMFFM), which enhances and fuses RGB and depth features with each other. Finally, we add edge prediction as an auxiliary task and propose an edge enhancement module (EEM) to generate sharper contours. Extensive experiments demonstrate that our method achieves superior performance over 14 state-of-the-art RGB-D methods on six public datasets. Our code will be released at \href{https://github.com/lc012463/TANet}{https://github.com/lc012463/TANet}.

% \href{}{http}.\href{github}{https://github.com/lc012463/TANet}

\end{abstract}


\section{Introduction}
Salient object detection (SOD) aims to model the mechanism of human visual attention and identify the most significant objects or regions in images of various visual scenes~\cite{1}. As a technology for preprocessing, it plays a fundamental and important role in many computer vision tasks~\cite{2} such as image retrieval~\cite{3}, target tracking~\cite{4}, image/video segmentation~\cite{5, 6}, person re-identification~\cite{7}, and image understanding~\cite{8}.

Although a large number of purely CNN-based methods~\cite{9, 10, 11, 12, 13} achieve quite promising results on the RGB SOD task, accurate identification and localization of salient objects in many challenging scenarios is still difficult~\cite{14}. With the wide availability of depth sensors in smart devices~\cite{10}, depth data have been introduced to provide geometric and spatial information to improve SOD performance. The depth cue has the preponderance of discriminative power in location and spatial structure~\cite{15}, which has been proven beneficial for accurate saliency prediction.

A great number of methods~\cite{9, 16, 17, 18, 19} for the RGB-D SOD task are CNN-based and leverage the symmetric two-stream structure to extract features from RGB and depth data separately. However, the inherent local connectivity of the convolution operation limits the performance of CNN-based methods~\cite{20}. And there is another level of ambiguity that is often neglected: if RGB and depth data are necessary to fit into the same network~\cite{21}. To be specific, the symmetric two-stream CNN-based methods have two fundamental limitations:

The first is that CNN-based models are ineffective in extracting global semantic information. In the RGB-D SOD task, long-range context and contrast information play a key role in identifying and locating salient objects~\cite{20}. Some researchers intend to improve the receptive field of CNN by using dilated convolutions~\cite{22}. However, due to the inherent limitations of the sliding window feature extraction method, CNN-based models still struggle to learn global semantic information. Recently, Transformer-based methods have outperformed existing state-of-the-art CNN-based methods in various computer vision tasks~\cite{23, 24, 25, 26} due to their powerful ability to extract features and model long-range dependencies, allowing Transformers to learn global semantic information effectively.

The second is that the symmetric two-stream structure ignores the inherent differences between modalities. Most existing RGB-D methods use the symmetric two-stream structure to extract RGB features and depth features separately~\cite{21, 27, 28}. However, RGB data have more semantic information for identifying and locating salient objects, such as color, texture, and limited location, while grayscale depth data contain more spatial structure information for refining the boundaries of salient objects, such as outlines and 3D layout~\cite{21}. Therefore, the symmetric two-stream structure may ignore the inherent differences between RGB and depth data, leading to inaccurate predictions.

Based on the above investigation, we strive to take a further step towards the goal of accurate saliency detection with a Transformer-based asymmetric network (TANet). In the RGB channel, we exploit the powerful feature extraction and long-range dependency modeling capabilities of the Transformer to extract global semantic information; in the depth channel, we design a lightweight CNN-based backbone (LWDepthNet), which does not require pre-training to effectively extract spatial structure information and better adapt to computation-intensive scenarios. In this way, the asymmetric hybrid encoder (AHE) can reduce the number of parameters in the model without sacrificing performance. To effectively integrate the information obtained from two different modalities, we design a cross-modal feature fusion module (CMFFM). It includes a depth feature enhancement block (DFEB), an RGB feature enhancement block (RFEB), and a fusion block. The DFEB enhances depth features by extracting semantic information from the RGB data, whereas the RFEB enhances RGB features by extracting spatial structure information from the depth data. After that, the enhanced features are fused. During the decoding process, the shallow features exhibit detailed boundary information and also introduce some background noise. Therefore, we add edge prediction as an auxiliary task and propose an edge enhancement module (EEM). The extracted boundary information is then used to enhance the saliency information and improve the prediction quality of the network.

In summary, our contributions are three folds:
\begin{itemize}
\item A novel Transformer-based asymmetric network (TANet) is proposed for the RGB-D SOD task. It extracts discriminative features from the asymmetric hybrid encoder (AHE), absorbing the local advantages of CNN and the advantages of Transformer in modeling long-range dependencies.

\item A newly designed cross-modal feature fusion module (CMFFM) is proposed to optimize and fuse the features of the two modalities. The depth features are spatially aligned and channel re-calibrated using the global semantic information extracted from the RGB data, and then the spatial structure information extracted from the depth data is introduced into the RGB features to generate fusion features.

\item Extensive experiments demonstrate that our method achieves superior performance over 14 state-of-the-art RGB-D methods on six public datasets, and the proposed model is much lighter than other Transformer-based methods, achieving a real-time speed of 33.6 FPS.
\end{itemize}

\section{Related works}

\subsection{CNN-based methods}

Although many RGB SOD methods~\cite{9, 10, 11, 12, 13} have achieved quite promising results, accurately identifying and localizing salient objects in many challenging scenarios remains difficult. Lang et al.~\cite{29} and Qu et al.~\cite{30} pioneered the introduction of depth maps in SOD tasks, hoping to mimic the human 3D visual perception system and use the spatial information contained in the depth maps to improve the prediction of saliency maps.

How to combine the information of the two modalities is a key issue in the RGB-D SOD task. Li et al.~\cite{31} proposed to modulate RGB features with depth features as priors through a cross-modal feature modulation module. Zhao et al.~\cite{17} designed a gating unit to select the features extracted by the encoder at different stages of the two modalities. Chen et al.~\cite{16} proposed a new complementarity-aware module that learns pairwise modal complementary information using asymptotically approximated residual functions. Pang et al.~\cite{18} proposed a hierarchical dynamic filtering network that effectively utilizes cross-modal fusion information.

Another key issue of the RGB-D SOD task is how to better utilize the deep semantic information and shallow spatial structure information. Liu et al.~\cite{9} proposed a global guidance module to guide the extracted global semantic information into the decoder, avoiding the dilution of global semantic information. Zhao et al.~\cite{10} used the edge features of salient objects to enhance the ability of the network to localize salient objects and refine the boundaries. Liu et al.~\cite{32} proposed a stereo attention mechanism to adaptively fuse features of various scales. Zhao et al.~\cite{33} used channel attention and spatial attention mechanisms to filter and fuse features.

However, the classical symmetric two-stream structure and simple fusion strategy may ignore the inherent difference between the two modalities, resulting in inaccurate prediction. Asymmetric architectures have been adopted in some works to extract RGB and depth features. Zhang et al.~\cite{21} proposed an asymmetric two-stream network, designed a flow ladder module for the RGB channel to fully extract global semantic information, and proposed a depth attention module to fully extract local spatial information in the depth channel. Wang et al.~\cite{34} designed a lightweight feature extraction network for the depth channel and proposed a cross-modality long-range context information gathering module to fuse the semantic information between the two modalities. Piao et al.~\cite{35} proposed an asymmetric two-stream network based on knowledge distillation to reduce the number of parameters and computation of the model, so that the model can perform well in computation-intensive and memory-intensive scenarios.

\subsection{Transformer-based methods}

The success of Vision Transformer~\cite{23} in the field of image recognition has led to the widespread application of Transformer models in computer vision tasks. Various Transformer backbones following the hierarchical structure of VGG~\cite{36} and ResNet~\cite{37} have emerged successively, such as Swin Transformer~\cite{24}, PVT~\cite{25, 26} and P2T~\cite{38}. These Transformer-based methods have achieved state-of-the-art performance in a variety of computer vision tasks such as detection ~\cite{70} and segmentation~\cite{71}, demonstrating the great potential of the Transformer.

Some researchers have begun to study Transformation-based saliency object detection methods. Dosovitskiy et al.~\cite{39} observed that Transformers had lost their ability to represent features in local space and used the difficulty-aware learning method to improve their local space representation. Liu et al.~\cite{40} proposed a Transformer-based method that further exploit the global semantic information  extracted from RGB data. Qiu et al.~\cite{41} proposed ABiUNet, using CNN to enhance the spatial representation ability of Transformer. Liu et al.~\cite{42} proposed the triple Transformer module to enhance the representation of high-level features and utilize the depth purification module to enhance the depth features. Aiming at the cross-modal fusion problem of the Transformer-based methods, Wang et al.~\cite{43} proposed a global/local cross-modal fusion module to fully fuse the features of the two modalities, and Pang et al.~\cite{44} proposed a Transformer-based information propagation path that can automatically align RGB features and depth features.

However, all known Transformer-based RGB-D methods use two-stream encoders with a symmetric structure, such as SwinNet~\cite{45}, which uses a two-symmetric Swin Transformer~\cite{24} as the encoder. We note that RGB data usually contain more semantic information, while depth data usually contain more spatial structure information, and the symmetric structure may ignore the inherent differences between the two modalities, leading to inaccurate predictions. In addition, the number of parameters and computational complexity of Transformer-based methods are generally high due to the global modeling capability of the self-attention mechanism. For example, the computational cost of SwinNet~\cite{45} and VST~\cite{40} mainly exists in two symmetric Transformer-based encoders, accounting for about 87.3\% and 79.5\% of the model parameters, respectively.

\section{Our Method}

\begin{figure*}[!t]
% Use the relevant command to insert your figure file.
% For example, with the graphicx package use
  \includegraphics[width=1\textwidth]{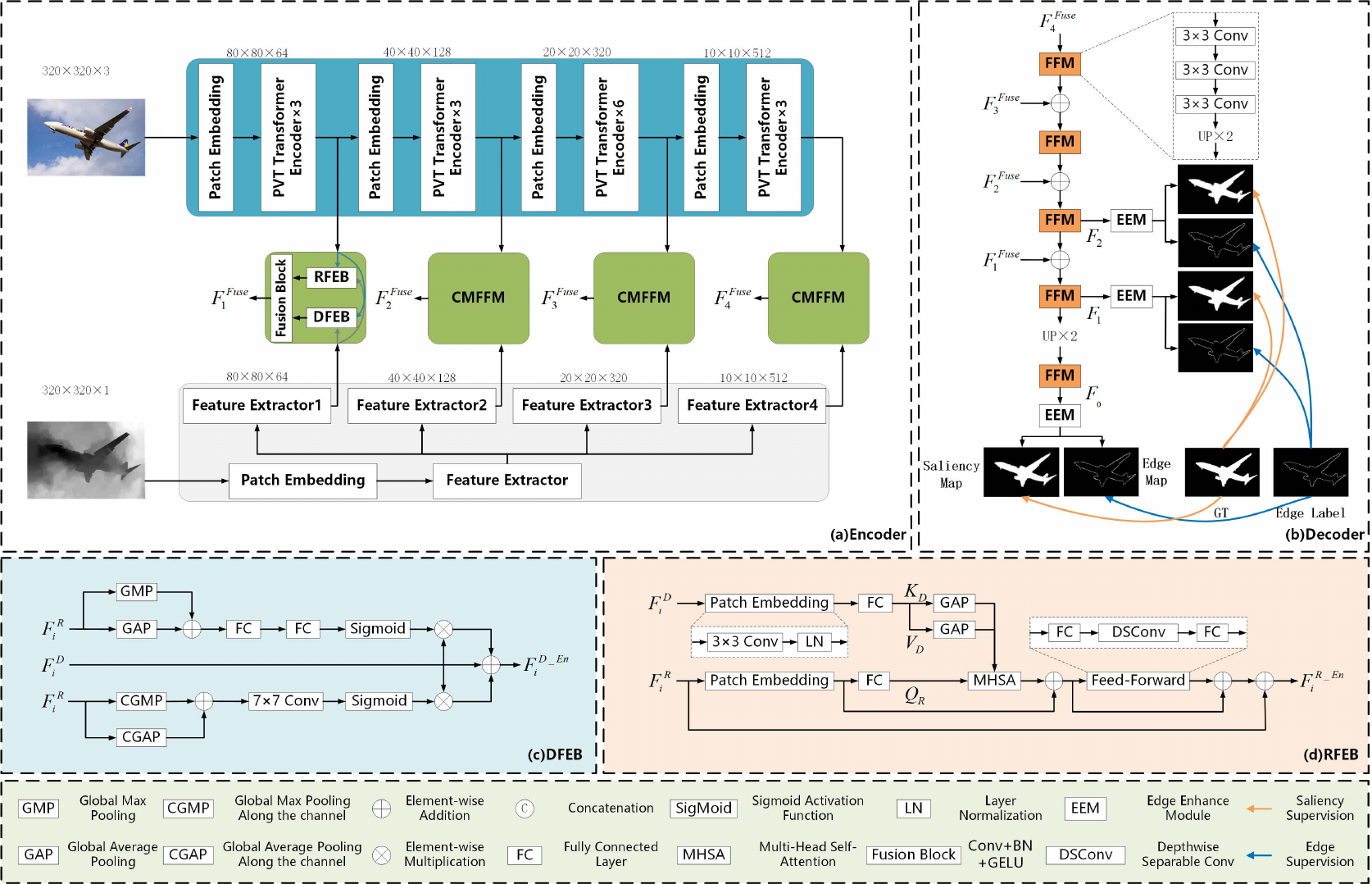}
% figure caption is below the figure
\caption{An overview of our proposed TANet. It consists of an asymmetric hybrid encoder, a cross-modal feature fusion module, an edge enhancement module, and a dual-task decoder. The multi-modal hierarchical features from the asymmetric hybrid encoder will be fed into the cross-modal feature fusion module to generate the fused features. Next, in the dual-task decoder, fused features are used to generate edge features and saliency features. Finally, we use the edge features to enhance saliency features and then generate saliency maps and edge maps.}
\label{fig:1}       % Give a unique label
\end{figure*}

\subsection{The overall architecture}\label{AA}

As shown in Fig.~\ref{fig:1}, we propose a Transformer-based asymmetric network (TANet). Inspired by~\cite{21, 34}, we design an asymmetric hybrid encoder (AHE), but unlike~\cite{21, 34}, AHE incorporates the advantages of Transformer in modeling long-range dependencies and the locality of CNN. To be more specific, in the RGB channel, we adopt a Transformer-based backbone (PVTv2) to extract global semantic information, while in the depth channel, we design a tailored lightweight CNN-based backbone (LWDepthNet) that does not require pre-training to extract local spatial information. The extracted features are then fed into the cross-modal feature fusion module (CMFFM). It performs spatial alignment and channel re-calibration of depth features using the global semantic information extracted from RGB data, and then introduces the local spatial information extracted from depth data into the RGB features to create fused features. Finally, we feed the fused features into a dual-task decoder and use the edge enhancement module (EEM) to improve the boundary quality of the predicted saliency maps.

\subsection{Asymmetric hybrid encoder}

Here, we want to explain that asymmetry mainly means: i) two encoder paths are based on different models (Transformer and CNN); ii) two encoder paths use different forms of feature extraction (hierarchical and parallel); iii) the goals they are responsible for are different (global semantic information extraction and local spatial information extraction).

\textbf{Encoder of RGB channel:} We use the Transformer-based PVTv2~\cite{26} as the encoder for the RGB channel to effectively extract the rich semantic information in the RGB data. PVTv2 is designed in a hierarchical manner and does not require positional encoding modules. This allows it to build multi-level features and accept inputs of any scale, as in the classical CNN backbones~\cite{31, 32}.

All stages of PVTv2 share a similar architecture, which consists of an overlapping patch embedding layer and several PVT blocks. Specifically, given an input image, PVTv2 splits it into small patches with the size of 4×4, using an overlapping patch embedding layer. Then, the flattened patches are fed into several PVT blocks. From the second stage, PVTv2 utilizes an overlapping patch embedding layer to shrink the feature map by a scale of 2 at the beginning of each stage, followed by some PVT blocks. Assume $\left\{F_{i}^{R} \mid i=1,2,3,4\right\}$ denote the output features of the four PVT stages from top to bottom, with scales of 1/4, 1/8, 1/16, and 1/32 and 64, 128, 320, and 512 channels, respectively. Please refer to the original paper~\cite{26} for more details.

% \begin{figure*}
% % Use the relevant command to insert your figure file.
% % For example, with the graphicx package use
%  \centering
%   \includegraphics[width=0.75\textwidth]{f2.pdf}
% % figure caption is below the figure
% \caption{Architecture of a standard PVT block. LSRA and CCF refer to linear spatial reduction attention and convloutional feed-forward modules, the purpose of which is to reduce the computational complexity of the network from square order to linear order.}
% \label{fig:2}       % Give a unique label
% \end{figure*}

\textbf{Encoder of depth channel:} Depth cues focus more on local spatial information than RGB data. In addition, the CNN model possesses translation invariance and locality, which have been proven beneficial for extracting local spatial information. Therefore, we argue that it is unnecessary to use a large Transformer-based complex network like PVTv2 to process depth data.
% Therefore, we consider it unnecessary to process depth data with a complex Transformer-based network as large as PVTv2.

\begin{figure}[!t]
    \includegraphics[width=0.5\textwidth]{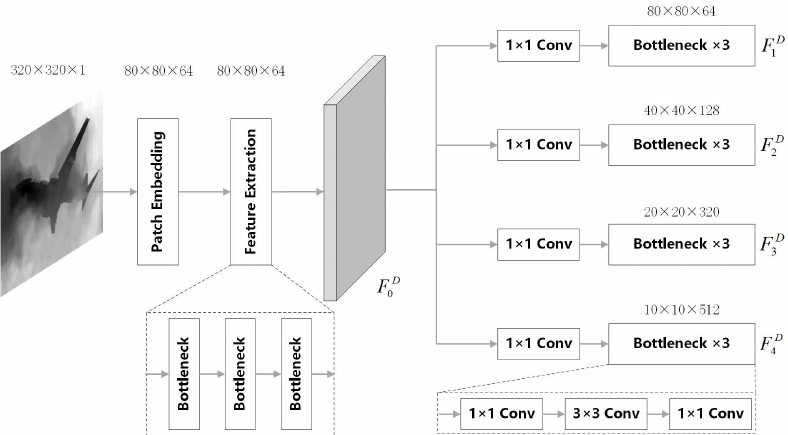}
    \caption{An overview of the proposed LWDepthNet.} 
    % To maximize the preservation of local spatial information in depth data, the features of different levels are generated in a parallel manner.
    \label{fig:3}      
\end{figure}

We use a parallel structure instead of a hierarchical structure to better preserve the spatial information of the depth data. And we employ convolutions with different strides instead of pooling layers for downsampling to prevent the loss of spatial structure information. Fig.~\ref{fig:3} illustrates the structural details of the proposed LWDepthNet. To begin with, in accordance with PVTv2, we employ an overlapping patch embedding layer to reduce the resolution of the input depth data to 1/4 of the input. The features are then encoded as base feature $F_{0}^{D}$ via a feature extraction module, which consists of three bottleneck layers~\cite{46} sequentially. Finally, we adjust the number of channels with a 1×1 convolutional layer and extract multi-scale features $\left\{F_{i}^{D} \mid i=1,2,3,4\right\}$ with four parallel feature extraction modules. Note that it is consistent with the shape of the multi-scale features output by the encoder of the RGB channel.

\subsection{Cross-modal feature fusion module}

We design a cross-modal feature fusion module (CMFFM) to integrate features from both modalities. It includes a depth feature enhancement block (DFEB), an RGB feature enhancement block (RFEB), and a fusion block.

\textbf{Depth feature enhancement block:} Since LWDepthNet is a lightweight CNN backbone of our own design, the extracted features may be quite different from the RGB features. To compensate for this difference, we propose a depth feature enhancement block that enhances depth features with semantic information extracted from RGB data. Specifically, we use a spatial attention module and a channel attention module to perform spatial alignment and channel re-calibration of the depth features.

A standard depth feature enhancement block is shown in Fig.~\ref{fig:1}~(c). Given the RGB features $F_{i}^{R}$ and depth features $F_{i}^{D}$ at a certain hierarchy $i \in\{1,2,3,4\}$, the enhanced depth feature $F_{i}^{D\_En}$ is computed as follows:

\begin{equation}
F_{i}^{D\_E n} = F_{i}^{D}+ SA(F_{i}^{R}, F_{i}^{D})+ CA(F_{i}^{R}, F_{i}^{D})
\end{equation}

where $SA(\cdot)$ and $CA(\cdot)$ represent the spatial attention mechanism and the channel attention mechanism, respectively. The computation procedure is as follows:

% \begin{equation}
% \begin{aligned}
% SA(F_{i}^{R}, F_{i}^{D}) = F_{i}^{D} \times &Sigmoid(FC(FC(GAP(F_{i}^{R}) \\
% &+ GMP(F_{i}^{R})))) \\
% CA(F_{i}^{R}, F_{i}^{D}) = F_{i}^{D} \times &Sigmoid(Conv_{7}(CGMP(F_{i}^{R}) \\
% &+ CGAP(F_{i}^{R})))
% \end{aligned}
% \end{equation}

\begin{equation}
\begin{aligned}
SA(F_{i}^{R}, F_{i}^{D}) = F_{i}^{D} \times &Sigmoid(FC(FC(GAP(F_{i}^{R}) \\
&+ GMP(F_{i}^{R})))) 
\end{aligned}
\end{equation}

\begin{equation}
\begin{aligned}
CA(F_{i}^{R}, F_{i}^{D}) = F_{i}^{D} \times &Sigmoid(Conv_{7}(CGMP(F_{i}^{R}) \\
&+ CGAP(F_{i}^{R})))
\end{aligned}
\end{equation}

where $GMP(\cdot)$ and $GAP(\cdot)$ denote a global maximum or average pooling operation. $CGMP(\cdot)$ and $CGAP(\cdot)$ denote a global maximum or average pooling operation along channel direction. $Conv_{7}(\cdot)$ represents the convolution operation with the kernel size 7×7, $FC(\cdot)$ represents the fully connected layer, and $Sigmoid(\cdot)$ denotes the sigmoid activation function.

\textbf{RGB feature enhancement block:} We propose an RGB feature enhancement block to introduce local spatial information extracted from depth data into the RGB features. 

We use a multi-head self-attention module to enhance RGB features. The specific details of the RGB feature enhancement block are shown in Fig.~\ref{fig:1}~(d). Given the RGB features $F_{i}^{R}$ and depth features $F_{i}^{D}$ at a certain hierarchy $i\in\{1,2,3,4\}$, we compute the enhanced RGB feature $F_{i}^{R\_En}$ as:

\begin{equation}
F_{i}^{R\_En}=F_{i}^{R}+CFFN(MHSA(Q_{R}, K_{D}, V_{D}))
\end{equation}

where $Q_{R}$ is the query extracted from the RGB features, $K_{D}$ and $V_{D}$ are the key and value extracted from the depth features. $MHSA(\cdot)$ and $CFFN(\cdot)$ represent the multi-head self-attention mechanism and convolutional feed-forward neural network modified by PVTV2, respectively. Please refer to the original paper~\cite{26} for more details.

After obtaining the enhanced features of both modalities, we fuse them with a fusion block. Suppose $\{F_{i}^{Fuse} \mid i=1,2,3,4\}$ denote the fused features, the procedure can be defined as:

\begin{equation}
F_{i}^{Fuse}=FusionB(F_{i}^{R\_En}+F_{i}^{D\_En})
\end{equation}

where $FusionB(\cdot)$ denotes the fusion block, which contains a convolution operation with the kernel size 3×3, a batch normalization layer, and a GELU activation function.

\subsection{Edge enhancement module}
\begin{figure*}[!t]
% Use the relevant command to insert your figure file.
% For example, with the graphicx package use
 \centering
  \includegraphics[width=0.75\textwidth]{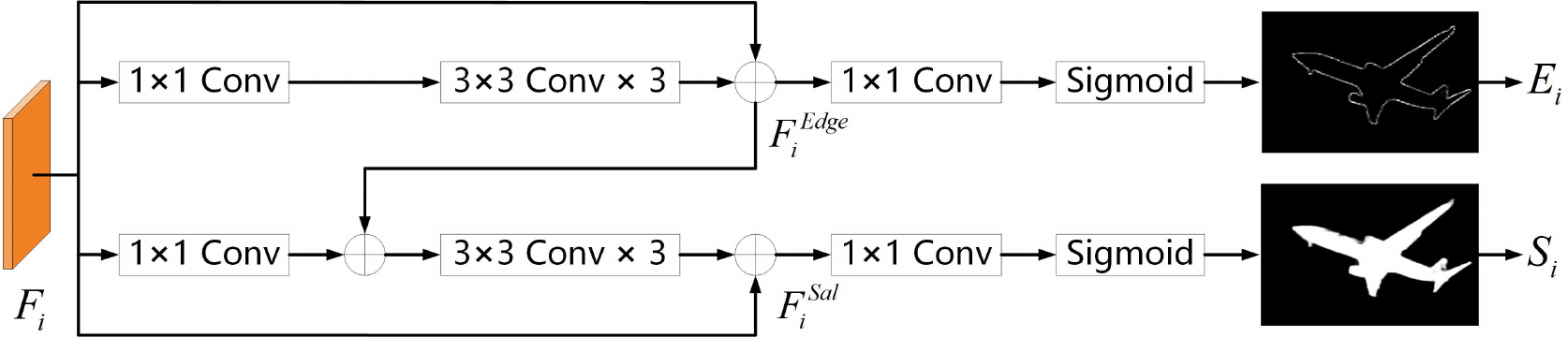}
% figure caption is below the figure
\caption{An overview of the proposed edge enhancement module.}
\label{fig:4}       % Give a unique label
\end{figure*}

To further improve the boundary quality of the predicted saliency maps, we propose an edge enhancement module (EEM). We add an extra edge prediction branch and utilize the resulting edge features to refine the boundaries of salient objects. The details of the proposed edge enhance module are illustrated in Fig.~\ref{fig:4}.

Considering that shallow features usually contain more boundary information, we only use features from the lower three layers for prediction. Given the features $F_{i}$ output by the decoder, we first compute the edge feature $F_{i}^{Edge}$ and then use it to enhance the salient feature $F_{i}^{Sal}$, where $i \in\{0,1,2\}$. It can be expressed as:

\begin{equation}
\begin{aligned}
&F_{i}^{Edge}=F_{i}+Conv(Conv_{1}(F_{i})) \\
&F_{i}^{Sal}=F_{i}+Conv(Conv_{1}(F_{i})+F_{i}^{Edge})
\end{aligned}
\end{equation}

% \begin{equation}
% F_{i}^{Edge}=F_{i}+Conv(Conv_{1}(F_{i}))
% \end{equation}

% \begin{equation}
% F_{i}^{Sal}=F_{i}+Conv(Conv_{1}(F_{i})+F_{i}^{Edge})
% \end{equation}

where $Conv(\cdot)$ denotes three convolution blocks. One convolution block consists of a convolution operation with the kernel size 3×3, a batch normalization layer, and a GELU activation function. $Conv_{1}(\cdot)$ represents the convolution operation with the kernel size 1×1.

Then, we use the extracted edge features $F_{i}^{Edge}$ and saliency features $F_{i}^{Sal}$ 
to predict edge maps $E_{i}$ and saliency maps $S_{i}$, where $i \in\{0,1,2\}$. It can be expressed as:

\begin{equation}
\begin{aligned}
&E_{i}=Sigmoid(Conv_{1}(F_{i}^{Edge})) \\
&S_{i}=Sigmoid(Conv_{1}(F_{i}^{Sal}))
\end{aligned}
\end{equation}

% \begin{equation}
% E_{i}=Sigmoid(Conv_{1}(F_{i}^{Edge}))
% \end{equation}

% \begin{equation}
% S_{i}=Sigmoid(Conv_{1}(F_{i}^{Sal}))
% \end{equation}

where $Conv_{1}(\cdot)$ represents the convolution operation with the kernel size 1×1, $Sigmoid(\cdot)$ denotes the sigmoid activation function.

\subsection{Loss function}
The loss function $L$ is defined as:
\begin{equation}
L=L_{E}(E_{i})+L_{S}(S_{i})
\end{equation}

where $L_{E}$ and $L_{S}$ denote edge loss and saliency loss, respectively. $E_{i}$ and $S_{i}$ refer to the predicted edge maps and saliency maps, respectively, where $i \in\{0,1,2\}$.

\textbf{Edge loss:} We use the edge ground truth supplied by~\cite{47} for supervision. The edge loss $L_{E}$ is defined as:

\begin{equation}
L_{E}=0.5  L_{BCE}(E_{1})+0.25 L_{BCE}(E_{2}) +0.25  L_{BCE}(E_{3})
\end{equation}

where $L_{BCE}$ denotes the binary cross-entropy loss, which can be formulated as:

\begin{equation}
L_{BCE}=\frac{\sum_{H} \sum_{W}(g \times \log (p)+(1-g) \times \log (1-p))}{H \times W}
\end{equation}

where $p$ denotes the predicted edge maps, $g$ denotes the corresponding ground-truth, $H$ and $W$ denote the high and width of the ground-truth respectively.

\textbf{Saliency loss:} The saliency loss function $L_{S}$ is defined as:

\begin{equation}
L_{S}=L_{IGL}(S_{1})+0.5  L_{I G L}(S_{2})+0.5  L_{I G L}(S_{3})
\end{equation}

where $L_{IGL}(\cdot)$ denotes the information-guided loss (IGL) proposed in~\cite{48}, which contains a BCE loss, a boundary loss, and an IOU loss. It can be formulated as:

\begin{equation}
L_{IGL}=L_{BCE}+\alpha  L_{B}+\beta  L_{IOU}
\end{equation}

where $\alpha$ and $\beta$ are the trade-off parameters, we simply set them as 1 and 0.7 respectively. Please refer to the original paper~\cite{48} for more details.

\section{Experiments}

\subsection{Datasets}
To validate the effectiveness of the proposed model, we perform our experiments on six public RGB-D datasets, including NJUD~\cite{49}, NLPR~\cite{50}, DUT-RGBD~\cite{51}, LFSD~\cite{52}, SIP~\cite{53}, and STERE~\cite{54}. For fair comparison, we use the same training dataset as in~\cite{44, 45}, which consists of 1,485 images from the NJUD dataset, 700 images from the NLPR dataset, and 800 images from the DUT-RGBD dataset. The remaining images from NJUD (500), NLPR (300) and DUT-RGBD (400), and the whole LFSD (100), SIP (929) and STERE (1,000), are used for testing.

\subsection{Evaluation metrics}
We adopt four evaluation metrics to evaluate the performance of saliency detection models: the precision-recall ($PR$) curve~\cite{55}; max F-measure score ($F_{\beta}$)~\cite{56}; mean absolute error ($MAE$)~\cite{57}; and structure-measure ($S_{m}$)~\cite{58}. The details of the four metrics are provided as follows:

\textbf{Precision-recall ($PR$):} Given a saliency map $S$, we convert it to a binary map $M$, and then we can compute the Precision and Recall by:
\begin{equation}
Precision=\frac{|M \cap G|}{|M|}, Recall=\frac{|M \cap G|}{|G|}
\end{equation}
where $G$ denotes the ground-truth. A popular strategy is to partition $S$ by using a set of thresholds (i.e., varying from 0 to 255). For each threshold, we calculate a pair of recall and precision scores, and then combine all the scores to obtain the PR curve.

\textbf{F-measure($F_{\beta}$):} The F-measure is used to comprehensively consider both precision and recall, and we can obtain the weighted harmonic mean by:
\begin{equation}
F_{\beta}=\frac{(1+\beta^{2}) Precision \times Recall}{\beta^{2} Precision + Recall}
\end{equation}
where $\beta^{2}$ is a trade-off parameter and it is set to 0.3 to emphasize the precision~\cite{56}.

\textbf{Mean Absolute Error ($MAE$):} The MAE is adopted to evaluate the average pixel-level relative error between the ground truth ($G$) and predicted saliency maps ($S$), which is defined by:

\begin{equation}
MAE=\frac{1}{W \times H} \sum_{i=1}^{W} \sum_{j=1}^{H}\left|S_{i, j}-G_{i, j}\right|
\end{equation}

where $W$ and $H$ denote the width and height of the map, respectively.

\textbf{Structure Measure($S_{m}$):} The S-measure is proposed to assess the structural similarity between the regional perception($S_{r}$) and object perception ($S_{o}$), which is defined by:
\begin{equation}
S=\alpha  S_{o}+(1-\alpha)  S_{r}
\end{equation}
where $a \in[0,1]$ is a trade-off parameter and it is set to 0.5 as default~\cite{58}.

\begin{table*}[!t]
\renewcommand\arraystretch{1.5}
\centering
\caption{Quantitative comparisons of F-measure($F_{\beta}$), MAE, and S-measure($S_{m}$) on 6 widely-used RGB-D datasets. The best three results are shown in \red{red}, \green{green}, \blue{blue} fonts respectively. From left to right: the latest CNN-based RGB-D methods and Transformer-based RGB-D methods.}
\label{tab:1}
% \begin{center}
% \resizebox{\textwidth}{!}{
\resizebox{\linewidth}{!}{ 
\begin{tabular}{cc|cccccccccc|cccc|c}
\toprule 
\multirow{3}{*}{} &\multirow{3}{*}{Metric} & HDFNet & D3Net & S2MA  & A2dele & DCF   & CDNet & RD3D  & DSA2F & JLDCF & HAINet & VST   & TriTrans & MTFormer & CAVER & \multirow{3}{*}{Ours} \\
                      &  & ECCV   & TNNLS & CVPR  & CVPR   & CVPR  & TIP   & AAAI  & CVPR  & TPAMI & TIP    & ICCV  & ACMM     & arXiv    & arXiv    &                       \\
                      &  & 2020   & 2020  & 2020  & 2020   & 2021  & 2021  & 2021  & 2021  & 2021  & 2021   & 2021  & 2021     & 2021     & 2022     &                       \\
\midrule
\multirow{3}{*}{\rotatebox{90}{NJUD}}          & $F_{\beta}$ ↑       & 0.922  & 0.910  & 0.899 & 0.890   & 0.916 & 0.926 & 0.923 & 0.917 & 0.904 & 0.921  & 0.920  & \textcolor[rgb]{0,1,0}{0.934}     & 0.923    & \textcolor[rgb]{0,0,1}{0.929}     & \textcolor[rgb]{1,0,0}{0.941}                  \\
                              & $MAE$ ↓      & 0.038  & 0.046 & 0.058 & 0.051  & 0.039 & 0.038 & 0.037 & 0.039 & 0.045 & 0.038  & 0.035 & \textcolor[rgb]{0,1,0}{0.030}     & \textcolor[rgb]{0,0,1}{0.032}    & 0.034    & \textcolor[rgb]{1,0,0}{0.027}                 \\
                              & $S_{m}$ ↑       & 0.908  & 0.900   & 0.887 & 0.871  & 0.904 & 0.916 & 0.916 & 0.904 & 0.893 & 0.909  & \textcolor[rgb]{0,0,1}{0.922} & 0.920     & 0.921    & \textcolor[rgb]{0,1,0}{0.925}    & \textcolor[rgb]{1,0,0}{0.927}                 \\
\hline
\multirow{3}{*}{\rotatebox{90}{NLPR}}          & $F_{\beta}$↑        & 0.927  & 0.907 & 0.910  & 0.898  & 0.917 & \blue{0.927} & \blue{0.927} & 0.915 & 0.925 & 0.918  & 0.920  & \green{0.930}     & 0.924    & 0.913    & \red{0.943}                 \\
                              & $MAE$ ↓      & 0.023  & 0.030  & 0.030  & 0.029  & 0.024 & 0.025 & \blue{0.022} & 0.024 & \blue{0.022} & 0.025  & 0.024 & \green{0.020}     & \green{0.020}     & 0.030     & \red{0.018}                 \\
                              & $S_{m}$↑       & 0.923  & 0.912 & 0.916 & 0.899  & 0.922 & 0.927 & \blue{ 0.930}  & 0.919 & 0.925 & 0.921  & \green{0.932} & 0.929    & \green{0.932}    & 0.924    & \red{0.935}                 \\
\hline
\multirow{3}{*}{\rotatebox{90}{DUT-R}}      & $F_{\beta}$ ↑       & 0.928  & 0.795 & 0.909 & 0.906  & 0.938 & 0.945 & 0.930  & 0.938 & 0.887 & 0.930   & \green{0.948} & \green{0.948}    & \blue{0.946}    & 0.938    & \red{0.960}                 \\
                              & $MAE$ ↓      & 0.041  & 0.096 & 0.042 & 0.043  & 0.030  & 0.030  & 0.038 & 0.030  & 0.070  & 0.038  & \green{0.024} & \blue{0.025}    & \green{0.024}    & 0.030     & \red{0.020}                  \\
                              &$S_{m}$ ↑       & 0.908  & 0.775 & 0.904 & 0.885  & 0.925 & 0.929 & 0.909 & 0.921 & 0.861 & 0.909  & \green{0.943} & 0.934    & \blue{0.937}    & 0.932    & \red{0.945}                 \\
\hline
\multirow{3}{*}{\rotatebox{90}{LFSD}}          & $F_{\beta}$ ↑       & 0.883  & 0.840  & 0.862 & 0.858  & 0.878 & 0.805 & 0.879 & \red{0.903} & 0.854 & 0.877  & 0.889 & \blue{0.890}     & 0.886    & -        & \green{0.892}                 \\
                              & $MAE$ ↓      & 0.077  & 0.095 & 0.095 & 0.077  & 0.071 & 0.062 & 0.073 & \red{0.055} & 0.081 & 0.079  & 0.061 & 0.066    & \blue{0.060}     & -        & \green{0.059}                 \\
                              & $S_{m}$ ↑       & 0.854  & 0.825 & 0.837 & 0.833  & 0.856 & 0.878 & 0.858 & \red{0.883} & 0.850  & 0.854  & \green{0.882} & 0.866    & \blue{0.880}     & -        & 0.875                 \\
\hline
\multirow{3}{*}{\rotatebox{90}{SIP}}           & $F_{\beta}$ ↑       & 0.910   & 0.881 & 0.891 & 0.856  & 0.900   & 0.908 & 0.906 & 0.891 & 0.904 & \green{0.916}  & \blue{0.915} & \green{0.916}    & 0.906    & 0.914    & \red{0.922}                 \\
                              & $MAE$ ↓      & 0.047  & 0.063 & 0.057 & 0.070   & 0.052 & 0.053 & 0.048 & 0.057 & 0.049 & 0.048  & \red{0.040}  & \blue{0.043}    & \red{0.040}     &     & \green{0.041}                \\
                              & $S_{m}$ ↑       & 0.886  & 0.860  & 0.872 & 0.829  & 0.874 & 0.879 & 0.885 & 0.862 & 0.880  & 0.886  & \red{0.904} & 0.886    & \blue{0.898}    & \green{0.901}    & 0.893                 \\
\hline
\multirow{3}{*}{\rotatebox{90}{STERE}}         & $F_{\beta}$ ↑       & 0.910   & 0.904 & 0.895 & 0.892  & 0.914 & 0.917 & 0.917 & 0.910  & 0.913 & \blue{0.919}  & 0.907 & \green{0.920}     & 0.906    & 0.906    & \red{0.934}                 \\
                              & $MAE$ ↓      & 0.041  & 0.046 & 0.051 & 0.045  & 0.037 & 0.038 & \blue{0.037} & 0.039 & 0.040  & 0.038  & 0.038 & \green{0.033}    & 0.039    & 0.041    & \red{0.027}                 \\
                              & $S_{m}$ ↑       & 0.900    & 0.899 & 0.891 & 0.879  & 0.906 & 0.909 & 0.911 & 0.898 & 0.903 & 0.909  & \green{0.913} & 0.908    & 0.906    & \blue{0.912}    & \red{0.923}     
                              \\
\bottomrule
\end{tabular}
}
% \end{center}
\end{table*}

\begin{figure*}[!t]
% Use the relevant command to insert your figure file.
% For example, with the graphicx package use
 \centering
  \includegraphics[width=0.8\textwidth]{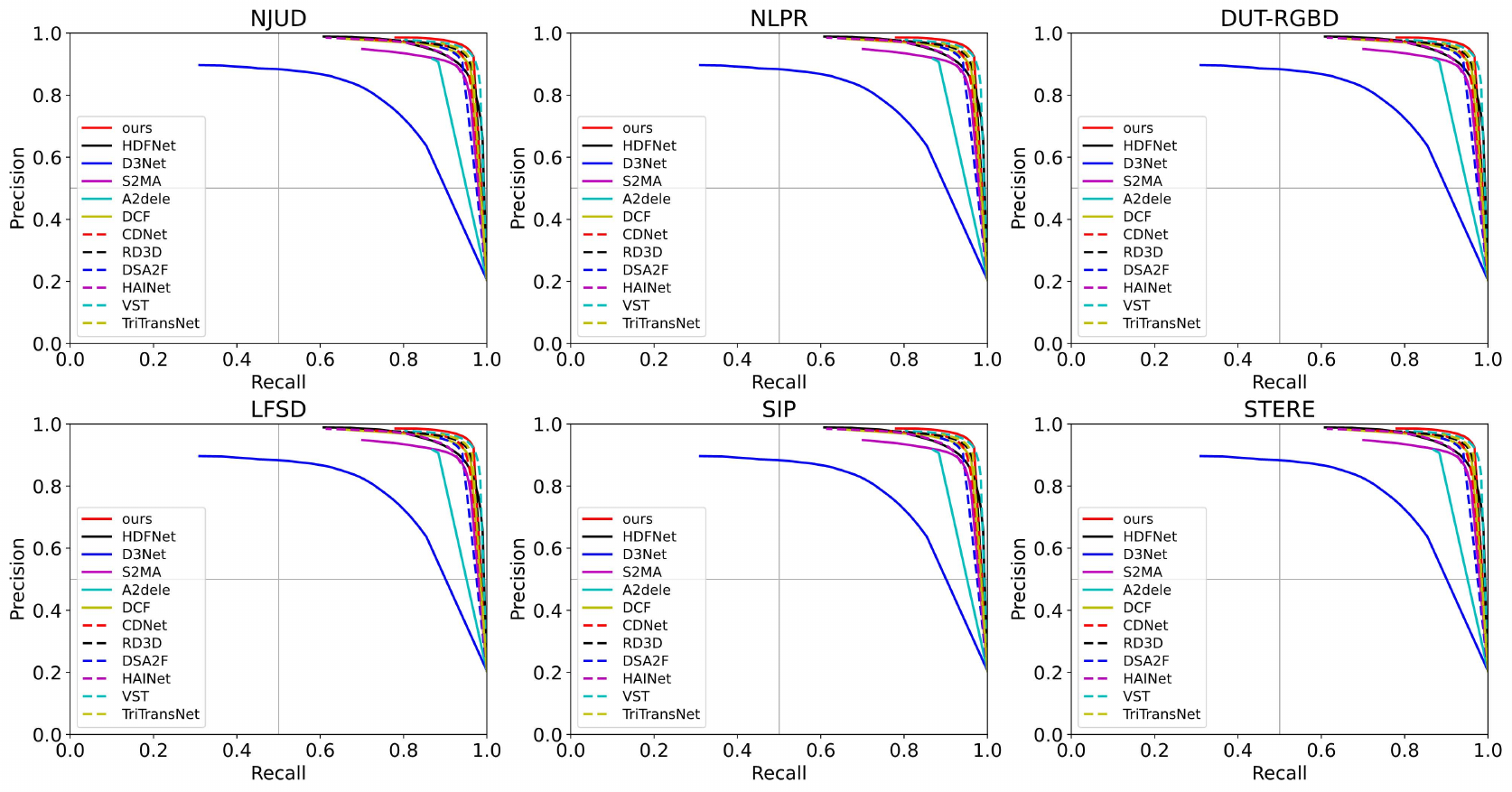}
% figure caption is below the figure
\caption{P-R curves comparison of different models on six RGB-D datasets. Our TANet represented by red solid line outperforms state-of-the-art models.}
\label{fig:5}       % Give a unique label
\end{figure*}

\begin{figure*}[!t]
% Use the relevant command to insert your figure file.
% For example, with the graphicx package us
 \centering
  \includegraphics[width=1\textwidth]{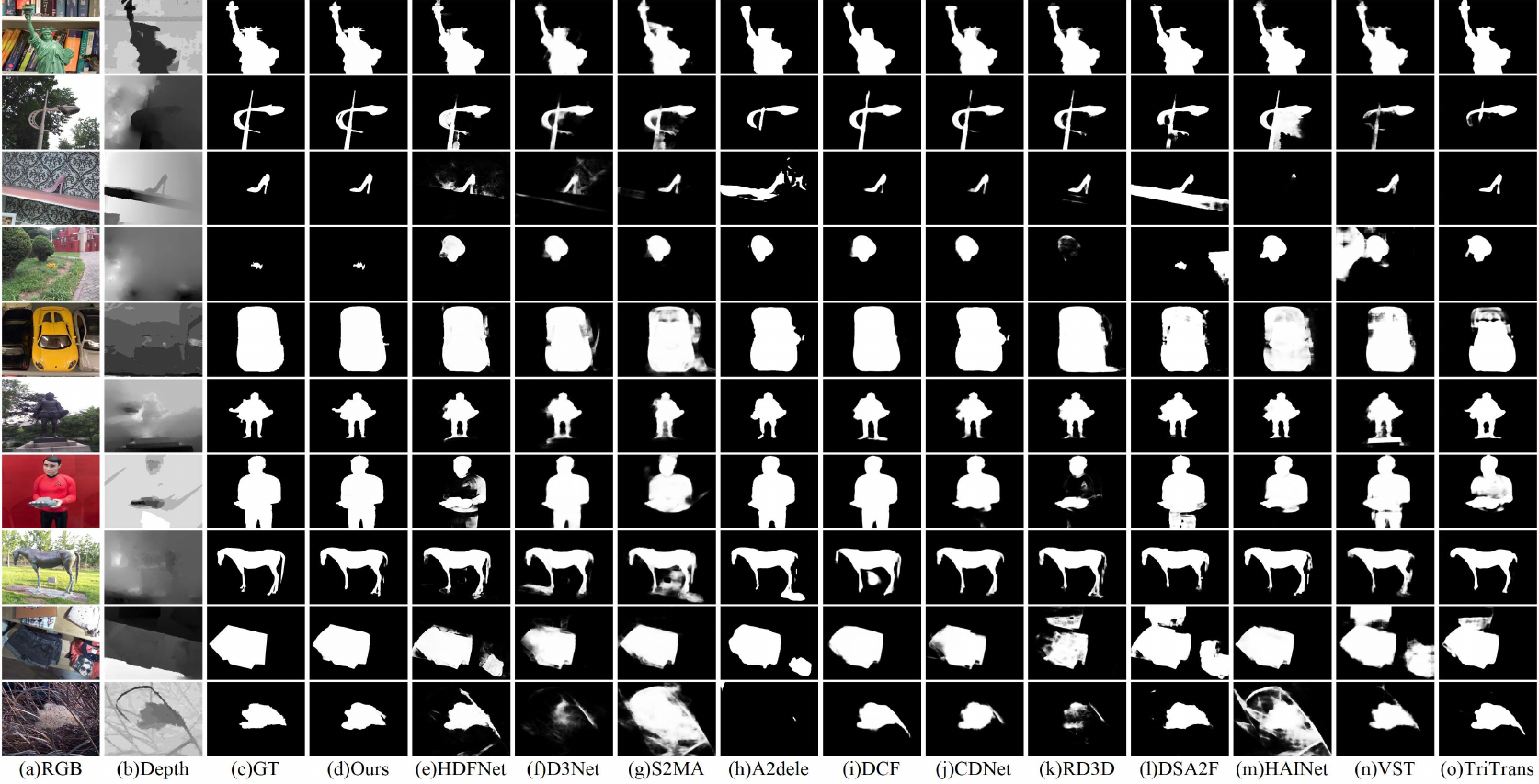}
% figure caption is below the figure
\caption{Visual comparison with state-of-the-art RGB-D models. Our TANet is outstanding in some challenging cases: edge and details ($1^{st} – 2^{nd}$ rows), small objects ($3^{rd} – 4^{th}$ rows), low quality depth map ($5^{th} – 6^{th}$ rows), big objects ($7^{th} – 8^{th}$ rows), and complex scenes ($9^{th} – 10^{th}$ rows).}
\label{fig:6}       % Give a unique label
\end{figure*}

\subsection{Implementation details}

The proposed model is implemented with the PyTorch library and trained on one GTX 3090 GPU with 24 GB memory. In both the training and testing phases, the input images are resized to 320×320, and the single-channel depth images are replicated as three-channel to meet the requirements. During the training process, the data augmentation method of random horizontal flipping and normalization are used to reduce overfitting. The parameters of the RGB channel’s backbone are initialized with the pretrained parameters of Pvt-v2-b2-linear~\cite{26} model. The rest of the parameters are initialized to PyTorch default settings. We employ the AdamW optimizer~\cite{59} to train our network with a momentum of 0.9, a mini-batch size of 8, and a weight decay of 1e-3 to optimize models. The initial learning rate is set to 1e-5 and decreases following the cosine strategy~\cite{60}. The model converges within 100 epochs, which takes nearly 6.5 hours.

\subsection{Comparisons with state-of-the-art methods}

In order to verify the effectiveness of our proposed model, we compared it with 14 state-of-the-art RGB-D SOD methods, including CNN-based HDFNet~\cite{61}, D3Net~\cite{53}, S2MA~\cite{62}, A2dele~\cite{63}, DCF~\cite{64}, CDNet~\cite{65}, RD3D~\cite{66}, DSA2F~\cite{67}, JLDCF~\cite{68}, HAINet~\cite{69}, and Transformer-based VST~\cite{40}, TriTransNet~\cite{42}, MutualFormer~\cite{43}, CAVER~\cite{44}. To ensure the fairness of the comparison results, the saliency maps of the evaluation are provided by the authors or generated by running source codes. And we evaluate them using the method presented in~\cite{50}.

\textbf{Quantitative Evaluation.} Table~\ref{tab:1} shows the quantitative comparison results of three evaluation metrics. As can be clearly observed from the table, on the NJUD, NLPR, DUT-RGBD, STERE datasets, the three evaluation metrics of our method are better than other methods, and on the LFSD and SIP datasets, our method also achieves second-best performance. Compared with the Transformer-based method VST~\cite{40} and TriTransNet~\cite{42}, the F-measure, MAE, and S-measure are improved about 1.7\%, 16.6\%, 0.7\%, and 0.9\%, 12.7\%, 1\% on average. In addition, we observe that Transformer-based models generally outperform CNN-based models, which demonstrates the strong potential of Transformers on salient object detection tasks. The comparison results on the PR curve are shown in Fig.~\ref{fig:5}. It can be seen that our method corresponds to a curve positioned more upward. The PR curve and qualitative results all verify the effectiveness and advantages of our proposed method in the RGB-D SOD task.

\textbf{Qualitative Evaluation.} To further explain the advantages of our method, we show some qualitative results in Fig.~\ref{fig:6}. It can be observed that our method is more competitive than other methods in many challenging scenarios: edge and details ($1^{st} – 2^{nd}$ rows), small objects ($3^{rd} – 4^{th}$ rows), low quality depth map ($5^{th} – 6^{th}$ rows), big objects ($7^{th} – 8^{th}$ rows), and complex scenes ($9^{th} – 10^{th}$ rows). These all indicate the robustness and effectiveness of our proposed method.

\subsection{Ablation studies}
% \subsubsection{Effectiveness of Proposed Module}

\begin{table*}[h]
\renewcommand\arraystretch{1.2}
\centering
\caption{Effectiveness analysis of the asymmetric structure. RGB Encoder: the backbone we used for the RGB channel; Depth Encoder: the backbone we used for the depth channel. The best results are shown in \textbf{bold}.}
\begin{center}
\begin{tabular}{c|cc|ccc|ccc|ccc}
\toprule 
\multirow{2}{*}{No.} & \multirow{2}{*}{RGB Encoder} & \multirow{2}{*}{Depth   Encoder} & \multicolumn{3}{c|}{NJUD} & \multicolumn{3}{c|}{NLPR} & \multicolumn{3}{c}{DUT-RGBD} \\
                    &                              &                                  & $F_{\beta}$ ↑   & $MAE$ ↓  & $S_{m}$ ↑   & $F_{\beta}$ ↑   & $MAE$ ↓  & $S_{m}$ ↑   & $F_{\beta}$ ↑     & $MAE$ ↓   & $S_{m}$ ↑    \\
\midrule
\#1                   & PVTv2                        & PVTv2                            & 0.939  & 0.028  & 0.923  & 0.938  & 0.018  & 0.932  & 0.959    & 0.020   & 0.944   \\
\#2                   & ResNet50                        & ResNet50                      & 0.927  & 0.036  & 0.911  & 0.930  & 0.022  & 0.925  & 0.949    & 0.027   & 0.932   \\
\#3                   & VGG16                        & VGG16                            & 0.928  & 0.036  & 0.911  & 0.925  & 0.022  & 0.921  & 0.941    & 0.032   & 0.920   \\
\#4                   & PVTv2                        & MobileNetv2                      & 0.941  & 0.027  & 0.926  & 0.938  & 0.019  & 0.931  & 0.960    & 0.020   & \textbf{0.947}   \\
\#5                   & PVTv2                        & ResNet50                         & 0.940  & 0.027  & 0.926  & 0.936  & 0.018  & 0.933  & 0.959    & 0.021   & 0.942   \\
\#6                   & PVTv2                        & VGG16                            & 0.940  & 0.027  & 0.927  & 0.938  & 0.019  & 0.933  & 0.958    & 0.020   & 0.944   \\
\#7                   & ResNet50                     & LWDepthNet                       & 0.928  & 0.036  & 0.909  & 0.921  & 0.906  & 0.023  & 0.943    & 0.030   & 0.923   \\
\#8                   & VGG16                        & LWDepthNet                       & 0.923  & 0.041  & 0.901  & 0.927  & 0.024  & 0.917  & 0.940    & 0.034   & 0.913   \\
\#9                   & Swin                         & LWDepthNet                       & 0.927  & 0.032  & 0.915  & 0.933  & 0.019  & 0.927  & 0.955    & 0.022   & 0.938   \\
\#10                   & P2T                          & LWDepthNet                       & 0.939  & 0.027  & 0.927  & 0.937  & 0.018  & 0.933  & 0.959    & 0.021   & 0.943   \\
\#11                   & PVTv2                        & LWDepthNet                       & \textbf{0.941}  & \textbf{0.027}  & \textbf{0.927}  & \textbf{0.943}  & \textbf{0.018}  & \textbf{0.935}  & \textbf{0.960}    & \textbf{0.020}   & 0.940  \\
\bottomrule
\end{tabular}
\label{tab:4}
\end{center}
\end{table*}

\begin{figure*}[h]
% Use the relevant command to insert your figure file.
% For example, with the graphicx package use
 \centering
  \includegraphics[width=1\textwidth]{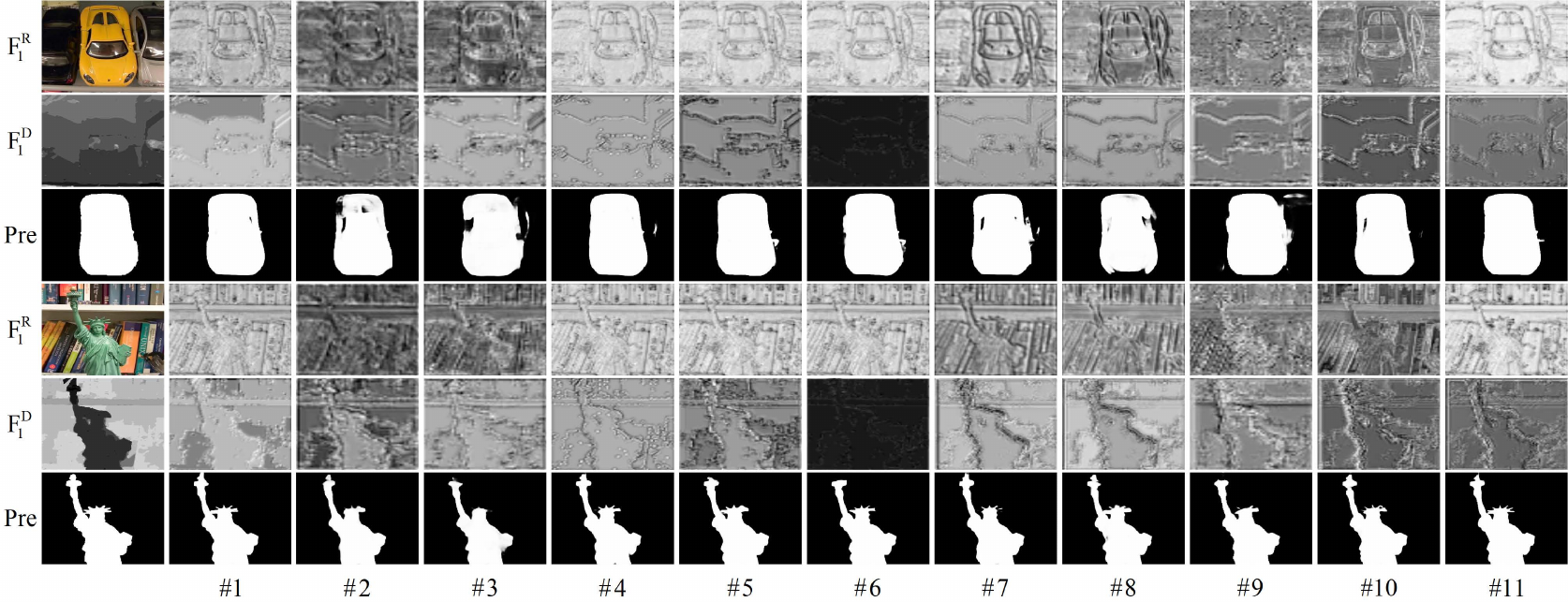}
% figure caption is below the figure
\caption{The visualization of the feature maps in the asymmetric structure. \#1-\#11 represent the asymmetric architectures indexed by the corresponding numbers in Tab~\ref{tab:4}. The first column is the RGB image, the depth image, and the ground truth. In other columns, there are features in layer 1, corresponding with the RGB features from the backbones $F_{1}^{R}$, the depth features from the backbones $F_{1}^{D}$, and the prediction saliency map $Pre$.}
\label{fig:8}       % Give a unique label
\end{figure*}

\textbf{Effect of asymmetric structure:} In Table~\ref{tab:4}, we replace the PVT and LWDepthNet backbones with some CNN backbones (e.g., VGG16~\cite{31}, ResNet50~\cite{32}, and MobileNetv2~\cite{46}) and Transformer backbones (e.g., Swin Transformer~\cite{24} and P2T~\cite{38}) to evaluate the effectiveness of our proposed asymmetric architecture. From the quantitative comparison results in Table~\ref{tab:4}, we can see that both symmetric and asymmetric two-stream structures, Tranformer-based methods (PVTv2, Swin, and P2T) generally achieve superior performance compared to pure CNN-based methods, which can be attributed to the powerful ability of Transformer to model long-range dependencies. Furthermore, neither Transformer-based nor CNN-based methods exhibit significant performance degradation when employing an asymmetric structure, demonstrating the effectiveness of our proposed asymmetric structure. It should be particularly noted that our proposed backbone of depth channel (LWDepthNet) achieves the best performance compared with other classical CNN backbones (see Table~\ref{tab:4} \#4-\#6 and \#11), proving the effectiveness of LWDepthNet.

In accordance with Table~\ref{tab:4}, we also perform some visual comparisons for different backbone combinations, as shown in Fig.~\ref{fig:8}. It can be found that there are indeed inherent differences in the features of the two modalities (see the $1^{st}–2^{nd}$ and $4^{th}–5^{th}$ rows). The features of the RGB channel contain more semantic information, while the features of the depth channel pay more attention to spatial structure information such as the outline of the salient object. In addition, it can be seen clearly that the RGB features extracted by PVTv2 contain less noise (see the $1^{st}$ and $4^{th}$ rows of \#4-\#6 and \#11) and the depth features extracted by LWDepthNet highlight the boundaries of salient objects (see the $2^{nd}$ and $5^{th}$ rows of \#7-\#11). Therefore, our method can not only localize the right salient objects but also maintain their sharp boundaries (see the $3^{rd}$ and $6^{th}$ rows of \#11), demonstrating the effectiveness of our proposed asymmetric structure.

\begin{table*}[h]
\renewcommand\arraystretch{1.2}
\centering
\caption{Effectiveness analysis of the modules in our proposed method, including CMFFM and EEM. The best results are shown in \textbf{bold}.}
\begin{center}
\begin{tabular}{c|ccc|ccc|ccc|ccc}
\toprule 
\multirow{2}{*}{No.} & \multirow{2}{*}{Baseline} & \multirow{2}{*}{CMFFM} & \multirow{2}{*}{EEM} & \multicolumn{3}{c|}{NJUD} & \multicolumn{3}{c|}{NLPR} & \multicolumn{3}{c}{DUT-RGBD} \\
                    &                           &                        &                      & $F_{\beta}$ ↑   & $MAE$ ↓  & $S_{m}$ ↑   & $F_{\beta}$ ↑   & $MAE$ ↓   & $S_{m}$ ↑  & $F_{\beta}$ ↑     & $MAE$ ↓   & $S_{m}$ ↑    \\
\midrule
\#1                   & \checkmark                         &                        &                      & 0.935  & 0.029  & 0.922  & 0.932  & 0.020  & 0.930 & 0.956    & 0.022   & 0.937   \\
\#2                   & \checkmark                         & \checkmark                      &                      & 0.939  & 0.028  & 0.922  & 0.937  & 0.018   & 0.933 & 0.959    & 0.020   & 0.940   \\
\#3                   & \checkmark                         &                        & \checkmark                    & 0.937  & 0.031  & 0.924  & 0.935  & 0.021   & 0.931 & 0.960    & 0.021   & 0.939   \\
\#4                   & \checkmark                         & \checkmark                      & \checkmark                    & \textbf{0.941}  & \textbf{0.027}  & \textbf{0.927}  & \textbf{0.943}  & \textbf{0.018}   & \textbf{0.935} & \textbf{0.960}    & \textbf{0.020}   & \textbf{0.940}  \\
\bottomrule
\end{tabular}
\label{tab:2}
\end{center}
\end{table*}

\begin{table*}[h]
\renewcommand\arraystretch{1.2}
\centering
\caption{Comparisons of different variants of CMFFM. w/o RFEB: removing RFEB, w/o DFEB: removing DFEB, w/o RFEB + w/o DFEB: removing both of RFEB and DFEB, w/o Fusion Block: removing the fusion block. The best results are shown in \textbf{bold}.}
\begin{center}
\begin{tabular}{c|c|ccc|ccc|ccc}
\toprule 
\multirow{2}{*}{No.} & \multirow{2}{*}{CMFFM} & \multicolumn{3}{c|}{NJUD} & \multicolumn{3}{c|}{NLPR} & \multicolumn{3}{c}{DUT-RGBD} \\
                    &                        & $F_{\beta}$ ↑   & $MAE$ ↓  & $S_{m}$ ↑   & $F_{\beta}$ ↑   & $MAE$ ↓  & $S_{m}$ ↑   & $F_{\beta}$ ↑     & $MAE$ ↓   & $S_{m}$ ↑    \\
\midrule
\#1                   & w/o RFEB               & 0.938  & 0.029  & 0.922  & 0.932  & 0.020  & 0.927  & 0.957    & 0.021   & 0.938   \\
\#2                   & w/o DFEB               & 0.939  & 0.028  & 0.926  & 0.935  & 0.020  & 0.931  & 0.957    & 0.020   & 0.939   \\
\#3                   & w/o RFEB + w/o DFEB    & 0.937  & 0.028  & 0.926  & 0.932  & 0.020  & 0.931  & 0.954    & 0.023   & 0.937  \\
\#4                   & w/o Fusion Block              & 0.939  & 0.029  & 0.925  & 0.938  & 0.019  & 0.932  & 0.960    & 0.022   & 0.940   \\
\#5                   & Ours                   & \textbf{0.941}  & \textbf{0.027}  & \textbf{0.927}  & \textbf{0.943}  & \textbf{0.018}  & \textbf{0.935}  & \textbf{0.960}    & \textbf{0.020}   & \textbf{0.940} \\
\bottomrule
\end{tabular}
\label{tab:3}
\end{center}
\end{table*}

\begin{figure*}[h]
% Use the relevant command to insert your figure file.
% For example, with the graphicx package us
 \centering
  \includegraphics[width=0.8\textwidth]{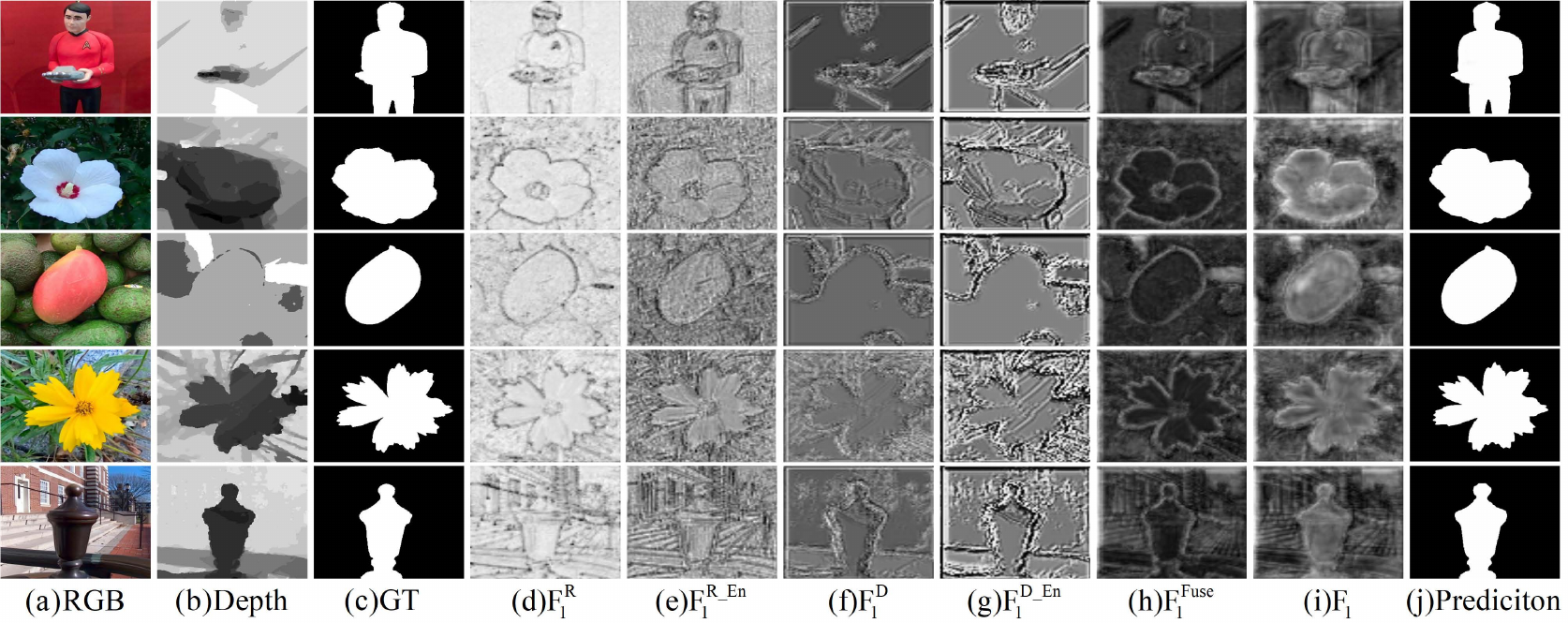}
% figure caption is below the figure
\caption{Visualization of features in CMFFM. From (a) to (c), there are the RGB image, depth image, and ground truth. (j) is the prediction saliency map. (d) to (i) are the features of level 1, corresponding with the RGB features from the backbones $F_{1}^{R}$, the RGB features through RFEB $F_{1}^{R \_ En}$, the depth features from the backbones $F_{1}^{D}$, the depth features through DFEB $F_{1}^{D \_ En}$, the fused features$F_{1}^{Fuse}$, and the features from the decoder $F_{1}$.}
\label{fig:7}       % Give a unique label
\end{figure*}

\begin{table*}[h]
\renewcommand\arraystretch{1.2}
\centering
\caption{Complexity comparisons on two datasets. The best results are shown in \textbf{bold}.}
\begin{center}
\begin{tabular}{c|c|c|c|ccc|ccc}
\toprule 
\multirow{2}{*}{No.} & \multirow{2}{*}{Methods} & \multirow{2}{*}{$Size$ ↓} & \multirow{2}{*}{$FPS$ ↑} & \multicolumn{3}{c|}{NJUD} & \multicolumn{3}{c}{NLPR} \\
                    &                          &                         &                       & $F_{\beta}$ ↑   & $MAE$ ↓  & $S_{m}$ ↑   & $F_{\beta}$ ↑   & $MAE$ ↓  & $S_{m}$ ↑   \\
    \midrule
\#1                   & VST                      & 80.4   MB               & 24.0                  & 0.920  & 0.035  & 0.922  & 0.920  & 0.024  & 0.932  \\
\#2                   & TriTransNet              & 139.5   MB              & 22.4                  & 0.934  & 0.030  & 0.920  & 0.930  & 0.020  & 0.929  \\
\#3                   & SwinNet                  & 198.7   MB              & 18.6                  & \textbf{0.943}  & 0.027  & \textbf{0.935}  & 0.941  & 0.018  & \textbf{0.941}  \\
\#4                   & PVTv2+PVTv2              & 66.0   MB               & 25.6                  & 0.939  & 0.028  & 0.923  & 0.938  & 0.018  & 0.932  \\
\#5                   & PVTv2+MobileNetv2        & 47.6   MB               & 33.3                  & 0.941  & 0.027  & 0.926  & 0.938  & 0.019  & 0.931  \\
\#6                   & PVTv2+ResNet50           & 69.0   MB               & 30.2                  & 0.940   & 0.027  & 0.926  & 0.936  & 0.018  & 0.933  \\
\#7                   & Swin+LWDepthNet          & 53.5 MB                 & 29.9                 & 0.927  & 0.032  & 0.915  & 0.933  & 0.019  & 0.927  \\
\#8                   & P2T+LWDepthNet           & 47.9 MB                 & 22.6                  & 0.939  & 0.027  & 0.927  & 0.937  & 0.018  & 0.933  \\
\#9                   & Ours                     & \textbf{47.5}   MB               & \textbf{33.6}                  & 0.941  & \textbf{0.027}   & 0.927  & \textbf{0.943}  & \textbf{0.018}  & 0.935 \\
\bottomrule
\end{tabular}
\label{tab:5}
\end{center}
\end{table*}

\textbf{Effectiveness of Proposed Module:} In order to verify the effectiveness of the modules in our proposed method, we conduct ablation studies on three public datasets. In Table~\ref{tab:2}, we quantitatively demonstrate the contribution of the cross-modal feature fusion module (CMFFM) and edge enhancement module (EEM) to our model. It can be seen that both CMFFM and EEM can improve the baseline performance, while the combination of these two modules leads to better performance. Moreover, we find that good results can be obtained with the baseline alone, which can be attributed to the powerful ability of Transformer to model long-range dependencies and the discriminative features extracted by the asymmetric hybrid encoder.

\textbf{Effectiveness of CMFFM:} We perform ablation studies on three public datasets to verify the impact of different variants of the cross-modal feature fusion module (CMFFM). The quantitative results are shown in Table~\ref{tab:3}. In Fig.~\ref{fig:7}, we also give some visual comparisons of the features in CMFFM. As shown in Table~\ref{tab:3}, the performance degradation caused by removing RFEB or DFEB proves that they are effective for enhancing RGB features and depth features. Meanwhile, it can be seen that the RFEB significantly improves the RGB features, sharpening the edges of salient objects (see Fig.~\ref{fig:7} (d) and (e)). The DFEB also improves the depth features, making it easier to distinguish the salient objects from their surrounding backgrounds (see Fig.~\ref{fig:7} (f) and (g)). This proves that our proposed RFEM and DFEM can effectively enhance the features of the two modalities. From Table~\ref{tab:3}, it can be observed that the fusion block can also improve the performance of CMFFM. In addition, we find that the fused features combine the advantages of the two modalities (see Fig.~\ref{fig:7} (h)), focusing not only on the whole of the salient object, but also on the boundary, proving the effectiveness of the CMFFM.

% Visualization of features in CMFFM.

\textbf{Complexity Evaluation:} We compare the model size and execution time of our method with the other eight Transformer-based models, as shown in Table~\ref{tab:5}. It can be seen that our method achieves the Top-1 FPS and the smallest model size. To be specific, the model size of our architecture is only 47.5MB and we achieve a high running speed of 33.6 frames per second (FPS). Compared to the best performing SwinNet~\cite{45}, our method achieves comparable performance with 76.1\% fewer parameters and 80.6\% higher FPS. Besides, our asymmetric structure achieves better performance than the symmetric structure (see Table~\ref{tab:5} \#4 and \#9) while minimizing the model size by 28\% and boosting the FPS by 31.5\%.

\section{Conclusion}
In this paper, we proposed a Transformer-based asymmetric network (TANet) to address the problem that the CNN-based models are ineffective in extracting global semantic information while the symmetric two-stream structures ignore the inherent differences between RGB and depth modalities. We employed the powerful feature extraction capabilities of Transformer to extract global semantic information from RGB data, and we designed a lightweight CNN backbone (LWDepthNet) without pre-training to extract the spatial structure information of the depth data. The proposed asymmetric hybrid encoder (AHE) effectively reduces the number of parameters and increases the speed without degrading performance. In addition, we proposed a CMFFM module to fuse the features of the two modalities and utilized the edge enhancement module (EEM) to generate sharper contours. Extensive experiments demonstrate that TANet can significantly improve the accuracy of the RGB-D SOD task. Moreover, our model size is only 47.5 MB and runs at a real-time speed of 33.6 FPS.

\section*{Acknowledgment}
This work is supported by the National Natural Science Foundation of China under Grant No. 62076058.
% [grant number aaaa]
% This work is supported by the National Natural Science Foundation of China under Grant No. 62076058.

% \bibliographystyle{IEEEtran}
% \small\bibliography{reference.bib}
% \small\bibliography{Manuscript.bbl}

\vspace{12pt}
\color{red}

\end{document}